\documentclass[runningheads,a4paper]{llncs}

\usepackage{amssymb}
\usepackage{amsmath}
\usepackage{graphicx}

\usepackage{url}
\urldef{\mailsa}\path|{alfred.hofmann, ursula.barth, ingrid.haas, frank.holzwarth,|
\urldef{\mailsb}\path|anna.kramer, leonie.kunz, christine.reiss, nicole.sator,|
\urldef{\mailsc}\path|erika.siebert-cole, peter.strasser, lncs}@springer.com|    
\newcommand{\keywords}[1]{\par\addvspace\baselineskip
\noindent\keywordname\enspace\ignorespaces#1}

\begin{document}

\mainmatter  

\title{Synthesis of Positron Emission Tomography (PET) Images via Multi-channel Generative Adversarial Networks (GANs)}

\titlerunning{Synthesis of Positron Emission Tomography (PET) Images via Multi-channel Generative Adversarial Networks (GANs)}

%
%
\author{Lei Bi\inst{1}  \and Jinman Kim\inst{1}  \and Ashnil Kumar\inst{1}  \and Dagan Feng\inst{1,2}  \and Michael Fulham\inst{1,3,4} }
\authorrunning{Lei Bi  \and Jinman Kim \and Ashnil Kumar  \and Dagan Feng  \and Michael Fulham }

\institute{School of Information Technologies, University of Sydney, Australia
\and Med-X Research Institute, Shanghai Jiao Tong University, China \and Department of Molecular Imaging, Royal Prince Alfred Hospital, Australia \and Sydney Medical School, University of Sydney, Australia}

%
%

\toctitle{Lecture Notes in Computer Science}
\tocauthor{Authors' Instructions}
\maketitle

\begin{abstract}
Positron emission tomography (PET) imaging is widely used for staging and monitoring treatment in a variety of cancers including the lymphomas and lung cancer. Recently, there has been a marked increase in the accuracy and robustness of machine learning methods and their application to computer-aided diagnosis (CAD) systems, e.g., the automated detection and quantification of abnormalities in medical images. Successful machine learning methods require large amounts of training data and hence, synthesis of PET images could play an important role in enhancing training data and ultimately improve the accuracy of PET-based CAD systems. Existing approaches such as atlas-based or methods that are based on simulated or physical phantoms have problems in synthesizing the low resolution and low signal-to-noise ratios inherent in PET images. In addition, these methods usually have limited capacity to produce a variety of synthetic PET images with large anatomical and functional differences. Hence, we propose a new method to synthesize PET data via multi-channel generative adversarial networks (M-GAN) to address these limitations. Our M-GAN approach, in contrast to the existing medical image synthetic methods that rely on using low-level features, has the ability to capture feature representations with a high-level of semantic information based on the adversarial learning concept. Our M-GAN is also able to take the input from the annotation (label) to synthesize regions of high uptake e.g., tumors and from the computed tomography (CT) images to constrain the appearance consistency based on the CT derived anatomical information in a single framework and output the synthetic PET images directly. Our experimental data from 50 lung cancer PET-CT studies show that our method provides more realistic PET images compared to conventional GAN methods. Further, the PET tumor detection model, trained with our synthetic PET data, performed competitively when compared to the detection model trained with real PET data (2.79\% lower in terms of recall). We suggest that our approach when used in combination with real and synthetic images, boosts the training data for machine learning methods.
\end{abstract}

\section{Introduction}

[18F]-Fluorodeoxyglucose (FDG) positron emission tomography (PET) is widely used for staging, and monitoring the response to treatment in a wide variety of cancers, including the lymphoma and lung cancer [1-3]. This is attributed to the ability of FDG PET to depict regions of increased glucose metabolism in sites of active tumor relative to normal tissues [1, 4]. Recently, advances in machine learning methods have been applied to medical computer-aided diagnosis (CAD) [5], where algorithms such as deep learning and pattern recognition, can provide automated detection of abnormalities in medical images [6-8]. Machine learning methods are dependent on the availability of large amounts of annotated data for training and for the derivation of learned models [7, 8]. There is, however, a scarcity of annotated training data for medical images which relates to the time involved in manual annotation and the confirmation of the imaging findings [9, 10]. Further, the training data need to encompass the wide variation in the imaging findings of a particular disease across a number of different patients. Hence effort has been directed in deriving other sources of training data such as ‘synthetic’ images. Early approaches used simulated, e.g., Monte Carlo approaches [24, 25] or physical phantoms that consisted of simplified anatomical structures [11]. Unfortunately, phantoms are unable to generate high-quality synthetic images and cannot simulate a wide variety of complex interactions, e.g., presence of the deformations introduced by disease. Other investigators used atlases [12] where different transformation maps were applied on the atlas with an intensity fusion technique to create new images. However, atlas based methods usually require many pre-/post-processing steps and a priori knowledge for tuning large amounts of transformation parameters, and thus limiting their ability to be widely adopted. Further, image registration that is used for creating the transformation maps affects the quality of the synthetic images.

\par

In this paper, we propose a new method to produce synthetic PET images using a multi-channel generative adversarial network (M-GAN). Our method exploits the state-of-the-art GAN image synthesis approach [13-16] with a novel adaptation for PET images and key improvements. The success of GAN is based on its ability to capture feature representations that contain a high-level of semantic information using the adversarial learning concept. A GAN has two competing neural networks, where the first neural network is trained to find an optimal mapping between the input data to the synthetic images, while the second neural network is trained to detect the generated synthetic images from the real images. Therefore, the optimal feature representation is acquired during the adversarial learning process. Although GANs have had great success in the generation of natural images, its application to PET images is not trivial. There are three main ways to conduct PET image synthesis with GAN: (1) PET-to-PET; (2) Label-to-PET; and (3) Computed tomography (CT)-to-PET. For PET-to-PET synthesis, it is challenging to create new variations of the input PET images, since the mapping from the input to the synthetic PET cannot be markedly different. Label-to-PET synthesis usually has limited constraints in synthesizing PET images, so the synthesized PET images can lack spatial and appearance consistency, e.g., the lung tumor appears outside the thorax. CT-to-PET synthesis is not usually able to synthesize high uptake regions e.g., tumors, since the high uptake regions may not be always visible as an abnormality on the CT images. Both PET-to-PET and CT-to-PET synthesis require new annotations for the new synthesized PET images for machine learning. Our proposition to address these limitations is a multi-channel GAN where we take the annotations (labels) to synthesize the high uptake regions and then the corresponding CT images to constrain the appearance consistency and output the synthetic PET images. The label is not necessary to be derived from the corresponding CT image, where user can draw any high uptake regions on the CT images which are going to be synthesized. The novelty of our method, compared to prior approaches, is as follows: (1) it harnesses high-level semantic information for effective PET image synthesis in an end-to-end manner that does not require pre-/post-processing or parameter tuning; (2) we propose a new multi-channel generative adversarial networks (M-GAN) for PET image synthesis. During training, M-GAN is capable of learning the integration from both CT and label to synthesize the high uptake and the anatomical background. During predication, M-GAN uses the label and the estimated synthetic PET images derived from CT to gradually improve the quality of the synthetic PET image; and (3) our synthetic PET images can be used to boost the training data for machine learning methods.

\section{Methods}

\subsection{Multi-channel Generative Adversarial Networks (M-GANs)}

\begin{figure}
\centering
\includegraphics[width=1.0\textwidth]{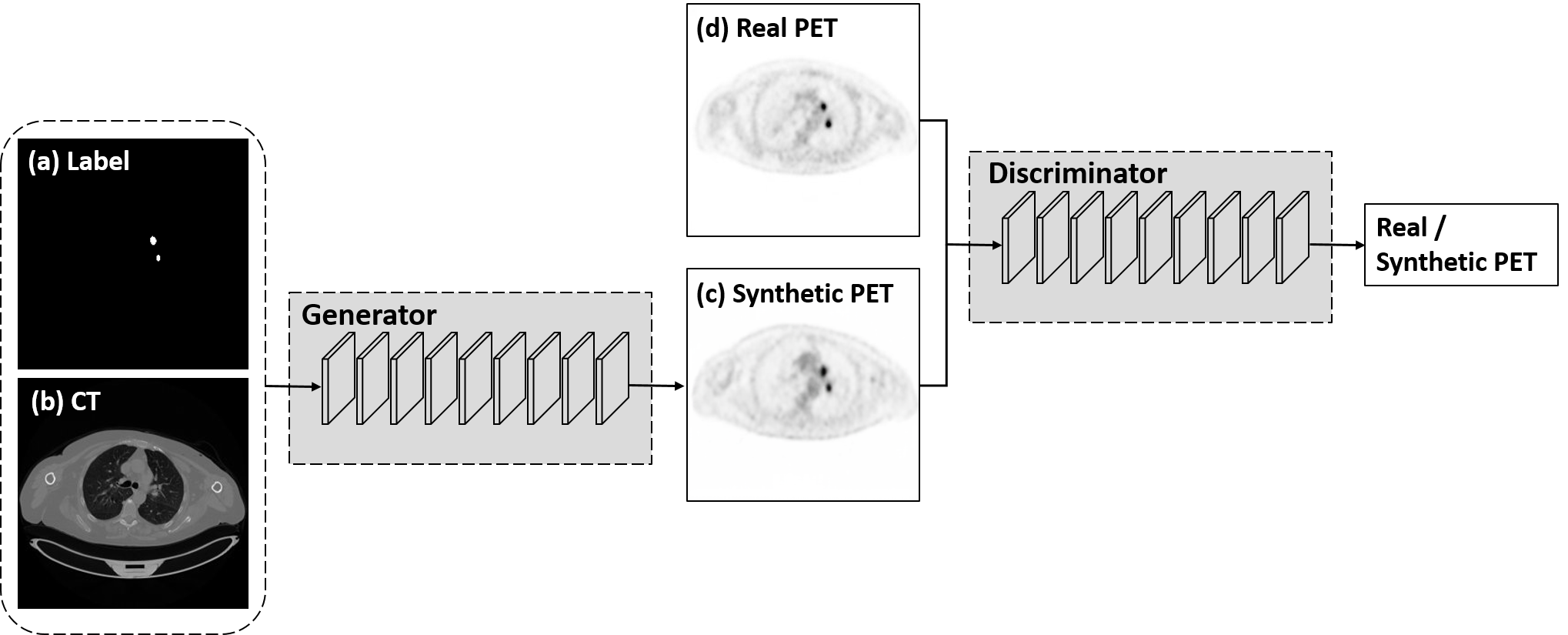}
\caption{Flow diagram of our M-GANs.}
\label{fig:example}
\end{figure}

GANs [13] have 2 main components: a generative model $G$ (the generator) that captures the data distribution and a discriminative model $D$  (the discriminator) that estimates the probability of a sample that came from the training data rather than $G$. The generator is trained to produce outputs that cannot be distinguished from the real data by the adversarially trained discriminator, while the discriminator was trained to detect the synthetic data created by the generator.

\par

Therefore, the overall objective is to minimize min-max loss function, which is defined as:

\begin{equation}
L(G,D)=\mathbb{E}_{x \sim p_{data(x)}} [logD(x)]+\mathbb{E}_{z \sim p_{z (z)}} [log(1-D(G(z)))]
\end{equation}

where $x$ is the real data and $z$ is the input random noise. $p_{data}$, $p_z$ represent the distribution of the real data and the input noise. $D(x)$ represents the probability that $x$ came from the real data while $G(z)$ represents the mapping to synthesize the real data. 

\par

For our M-GAN, we embed the label and the CT image for training and testing, as shown in Figure 1. During the training time, the generator takes input from the label and CT to learn a mapping to synthesize the real PET images. Then the synthesized PET images, together with the real PET images, enter into the discriminator for separation as:

\begin{multline}
L_{M-GAN}(G,D)=\mathbb{E}_{l,c,t \sim p_{data(l,c,t)}} [logD(l,c,t)]+ \\
\mathbb{E}_{{c \sim p_{c (c)}}, {l \sim p_{l (l)}}}[log(1-D(l,c,G(l,c)))]
\end{multline}

where $l$ is the label, $c$ the CT and $t$ is the PET image. The conceptual approach to train the M-GAN is to find an optimal setting $G^*$ that maximizes $D$ while minimizing $G$, which can be defined as:

\begin{equation}
G^*={arg \min }_{G} {\max }_{D} L_{M-GAN}(G,D)
\end{equation}

Based on the latest empirical data reported by van den Oord et al [14], we used $L1$ distance to encourage less blurring for the synthetic images during training. Therefore, the optimization process becomes:

\begin{equation}
G^*={arg \min }_{G} {\max }_{D} L_{M-GAN}(G,D)+\lambda \mathbb{E}_{{c \sim p_{c (c)}}, {l \sim p_{l (l)}}}[\Vert t-G(l,c)\Vert _1]
\end{equation}

where $\lambda$ is a hyper-parameter, which balances the contribution of the two terms and we set it to 100 empirically. We followed the published work Isola et al [15] and used a U-net [17] architecture for the generator $G$ and a five-layer convolutional networks for the discriminator $D$.

\subsection{Materials and Implementation Details}

Our dataset consisted of 50 PET-CT studies from 50 lung cancer patients provided by the Department of Molecular Imaging, Royal Prince Alfred (RPA) Hospital, Sydney, NSW, Australia. All studies were acquired on a 128-slice Siemens Biograph mCT scanner; each study had a CT volume and a PET volume. The reconstructed volumes had a PET resolution of 200$\times$200 pixels at 4.07mm\textsuperscript{2}, CT resolution of 512$\times$512 pixels at 0.98mm\textsuperscript{2} and slice thickness of 3mm. All data were de-identified. Each study contained between 1 to 7 tumors. Tumors were initially detected with a 40\% peak SUV (standardized uptake value) connected thresholding to detect ‘hot spots’. We used the findings from the clinical reports to make manual adjustments to ensure that the segmented tumors were accurate. The reports provided the location of the tumors and any involved lymph nodes in the thorax. All scans were read by an experienced clinician who has read 60,000 PET-CT studies. 

\par

To evaluate our approach we carried out experiments only on trans-axial slices that contained tumors and so analyzed 876 PET-CT slices from 50 patient studies. We randomly separated these slices into two groups, each containing 25 patient studies. We used the first group as the training and tested on the second group, and then reversed the roles of the groups. We ensured that no patient PET-CT slices were in both training and test groups. Our method took ~6 hours to train over 200 epochs with a 12GB Maxwell Titan X GPU on the Torch library [18].

\section{Evaluation}

\subsection{Experimental Results for PET Image Synthesis}

We compared our M-GAN to single channel variants: the LB-GAN (using labels) and the CT-GAN (using CTs). We used mean absolute error (MAE) and peak signal-to-noise ratio (PSNR) for evaluating the different methods [19]. MAE measures the average distance between each corresponding pixels of the synthetic and the real PET image. PSNR measures the ratio between the maximum possible intensity value and the mean squared error of the synthetic and the real PET image. The results are shown in Table 1 where the M-GAN had the best performance across all measurements with the lowest MAE and highest PSNR.

\begin{table}[]
\centering
\caption{Comparison of the different GAN approaches.}
\label{lab:2}
\def\arraystretch{1.2}
\begin{tabular}{cccccc}
\hline
                & \textbf{MAE}  & \textbf{PSNR}  \\ \hline
\textbf{LB-GAN} & 7.98          & 24.25          \\ \hline
\textbf{CT-GAN} & 4.77          & 26.65          \\ \hline
\textbf{M-GAN}  & \textbf{4.60} & \textbf{28.06} \\ \hline
\end{tabular}
\end{table}

\begin{figure}
\centering
\includegraphics[width=1.0\textwidth]{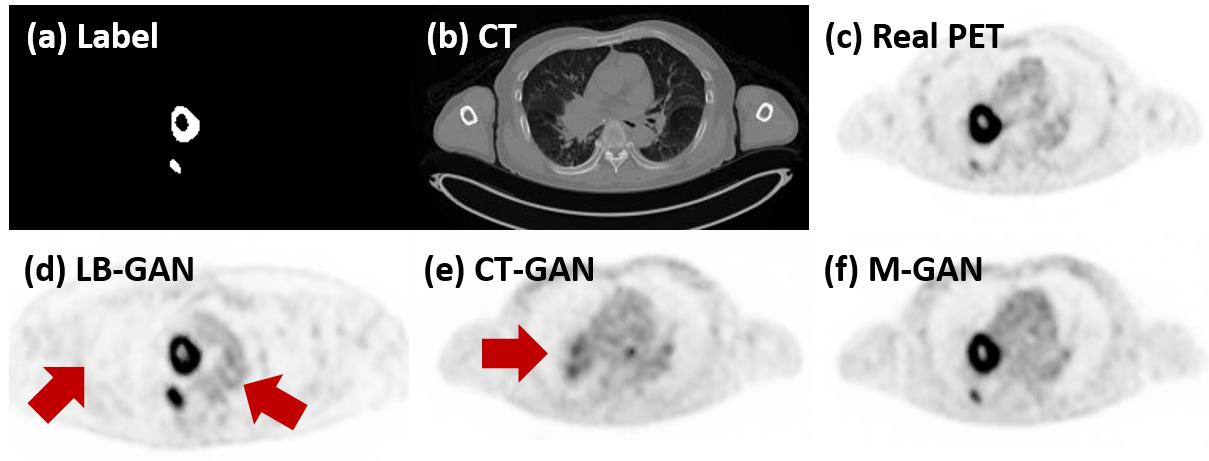}
\caption{Synthetic PET images. (a) label, (b) CT image, (c) real PET image, (d, e) synthetic PET images produced with only using (a) or (b), (f) our synthetic PET images with both (a) and (b) as the input.}
\label{fig:example}
\end{figure}

\subsection{Using Synthetic PET Images for Training}

In the second experiment, we analysed the synthetic PET images to determine their contribution to train a fully convolutional network (FCN – a widely used deep convolutional networks for object detection task [20-22]). We trained the FCN model with (i) LB-GAN, (ii) CT-GAN, (iii) M-GAN produced synthetic or (iv) real PET images. Then we applied the trained FCN model to detect tumors on real PET images (We used the first group to build the GAN model and the GAN model was applied on the second group to produce the synthetic PET images. After that, the synthetic PET images were used to build the FCN model. Finally the trained FCN model was tested on the first group with the real PET images for tumor detection. We reversed the roles of the two groups and applied the same procedures). Our evaluation was based on the overlap ratio between the detected tumor and the ground truth annotations [23]. A detected tumor with $>$50\% overlap with the annotated tumor (ground truth) was considered as true positive; additional detected tumor was considered as false positive. We regarded an annoted tumor that was not detected, or an overlap, smaller than 50\%, between the detected tumor and the annoted tumors as false negative. We measured the overall precision, recall and f-score.

\par

Table 2 shows the detection and segmentation performances. The results indicate that the M-GAN synthesized PET images performed competitively to the results produced from using real PET images for tumor detection.

\begin{table}[]
\centering
\caption{Comparision of FCN-based tumor detection performance, trained using synthetic or  real PET.}
\label{lab:2}
\def\arraystretch{1.4}
\begin{tabular}{c|ccc}
\hline
\textbf{Trained FCN with} & \textbf{Precision} & \textbf{Recall} & \textbf{F-score} \\ \hline
\textbf{LB-GAN PET}       & 76.42              & 44.06           & 55.90            \\ \hline
\textbf{CT-GAN PET}       & 36.89              & 3.69            & 6.71             \\ \hline
\textbf{M-GAN PET}        & 81.73              & 52.38           & 63.84            \\ \hline
\textbf{Real PET}         & 88.31              & 55.17           & 66.38            \\ \hline
\end{tabular}
\end{table}

\section{Discussion}

Table 1 indicates that the M-GAN is much closer to the real images when compared with other GAN variants, and achieved the lowest MAE score of 4.60 and a highest PSNR of 28.06. The best score in both MAE and PSNR can be used to indicate the construction of the most useful synthetic images. In general, LB-GAN may be employed to boost the training data. However, due to the lack of spatial and appearance constraints that could be derived from CT, LB-GAN usually result in poor anatomical definitions, as exemplified in Figure 2d, where the lung boundaries were missing and the mediastinum regions were synthesized wrongly.

\par

CT-GAN achieved competitive results in terms of MAE and PSNR (Table 1). However, its limitation is with its inability to reconstruct the lesions which are information that is only available in the label images (or PET), as exemplified in Figure 2e, where the two tumors were missing and one additional tumor was randomly appeared in the heart region from the synthetic images. The relative small differences between the proposed M-GAN method and CT-GAN method was due to the fact that tumor regions only occupy a small portion of the whole image and therefore, resulting less emphasis for the overall evaluation. In general, CT-GAN cannot synthesize the high uptake tumor regions, especially for the tumors adjacent to the mediastinum. This is further evidence in Table 2; CT-GAN synthesized PET images have inconsistent labeling of the tumors and resulting the trained FCN producing the lowest detection results.   

\par

In Table 2, the difference between the M-GAN and the detection results by using the real PET images demonstrate the advantages in integrating label to synthesize the tumors and the CT to constrain the appearance consistency in a single framework for training.

\section{Conclusion}

We propose a new M-GAN framework for synthesizing PET images by embedding a multi-channel input in a generative adversarial network and thereby enabling the learning of PET high uptake regions such as tumors and the spatial and appearance constraint from the CT data. Our preliminary results on 50 lung cancer PET-CT studies demonstrate that our method was much closer to the real PET images when compared to the conventional GAN approaches. More importantly, the PET tumor detection model trained with our synthetic PET images performed competitively to the same model trained with real PET images. In this work, we only evaluated the use of synthetic images to replace the original PET; in our future work, we will investigate novel approaches to optimally combine the real and synthetic images to boost the training data. We suggest that our framework can potentially boost the training data for machine learning algorithms that depends on large PET-CT data collection, and can also be extended to support other multi-modal data sets as PET-MRI synthesis.

\end{document}